\documentclass[letterpaper, 10 pt, conference]{ieeeconf}  

\IEEEoverridecommandlockouts                              

\usepackage{graphicx}
\usepackage{amsmath}
\usepackage{bbding} 
\usepackage{amssymb}
\usepackage[utf8]{inputenc}  
\usepackage{multirow}
\usepackage{comment}
\usepackage[ruled,vlined]{algorithm2e}
\usepackage{xcolor}

\title{\LARGE \bf
GeoLanG: Geometry-Aware Language-Guided Grasping with Unified RGB-D Multimodal Learning
}

\author{Rui Tang*, Guankun Wang*, Long Bai, Huxin Gao, Jiewen Lai, Chi Kit Ng, 
\\
Jiazheng Wang, Fan Zhang, Hongliang Ren}

\begin{document}

\maketitle
\thispagestyle{empty}
\pagestyle{empty}

\begin{abstract}
Language-guided grasping has emerged as a promising paradigm for enabling robots to identify and manipulate target objects through natural language instructions, yet it remains highly challenging in cluttered or occluded scenes. Existing methods often rely on multi-stage pipelines that separate object perception and grasping, which leads to limited cross-modal fusion, redundant computation, and poor generalization in cluttered, occluded, or low-texture scenes. To address these limitations, we propose GeoLanG, an end-to-end multi-task framework built upon the CLIP architecture that unifies visual and linguistic inputs into a shared representation space for robust semantic alignment and improved generalization. To enhance target discrimination under occlusion and low-texture conditions, we explore a more effective use of depth information through the Depth-guided Geometric Module (DGGM), which converts depth into explicit geometric priors and injects them into the attention mechanism without additional computational overhead.
In addition, we propose Adaptive Dense Channel Integration, which adaptively balances the contributions of multi-layer features to produce more discriminative and generalizable visual representations.
Extensive experiments on the OCID-VLG dataset, as well as in both simulation and real-world hardware, demonstrate that GeoLanG enables precise and robust language-guided grasping in complex, cluttered environments, paving the way toward more reliable multimodal robotic manipulation in real-world human-centric settings.
\end{abstract}

\setlength{\textfloatsep}{6pt}

\section{Introduciton}

\begin{figure}[!t]
\centering
\includegraphics[width=0.46\textwidth]{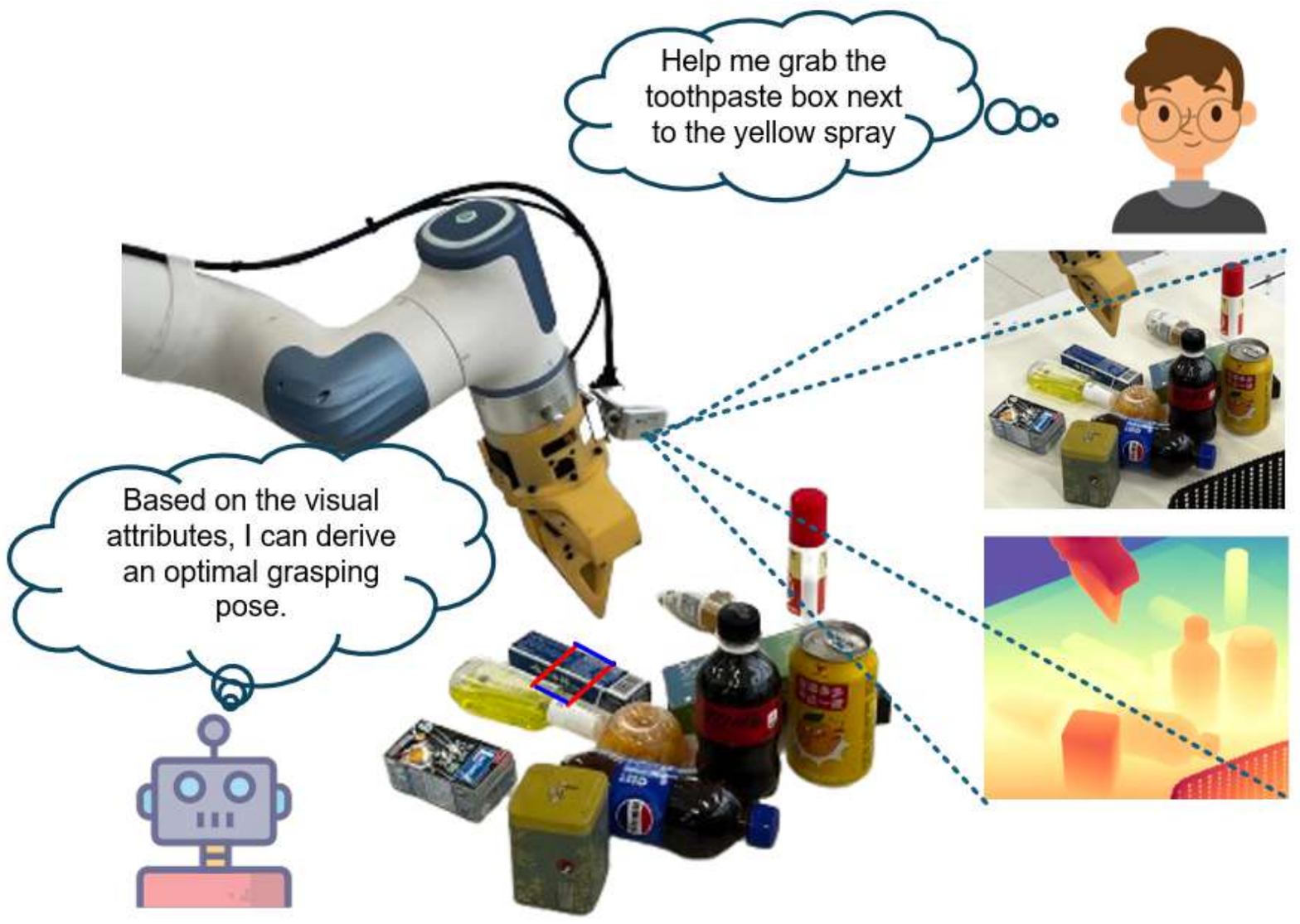}

\caption{Language-Guided Multimodal Perception for 6-DoF Robotic Grasping in Cluttered Environments.}
\label{fig_intro}
\end{figure}

In recent years, robots have been increasingly deployed in applications such as household assistance, elderly care, and warehouse logistics~\cite{hatori2018interactively,shridhar2020ingress,robotics11060127, ngendovla, hersh2015overcoming}. As shown in
Fig.~\ref{fig_intro}, a central challenge in this field is enabling robots to perform adaptive, explainable, and reliable grasping in open-world environments, where multiple objects may occlude or interfere with each other. Traditional RGB-based grasping pipelines typically separate object detection, segmentation, and grasp planning~\cite{mi13020293,tang2024optimizing}. While effective in controlled settings, these approaches often exhibit limited semantic understanding, poor generalization to cluttered or occluded scenes, and cumulative errors across stages. Language-guided grasping has emerged as a promising paradigm that uses natural language instructions to guide perception and planning. 
Contrastive language-image pretraining models, such as CLIP~\cite{radford2021learning}, provide strong capabilities for aligning visual and linguistic modalities. 
Recent approaches integrate CLIP with preprocessed object regions and Transformer encoders to construct multimodal fusion frameworks~\cite{wang2024video}. 
Although these methods improve semantic understanding, several limitations remain. 
First, many approaches rely on external object and grasp detectors, which are susceptible to cascading errors. 
Second, commonly adopted visual encoders such as CLIP-ResNet and CLIP-ViT each face intrinsic drawbacks: CLIP-ResNet, constrained by its static hierarchical structure, cannot dynamically model object scale variations and deformations, thereby limiting its representation of multi-scale and non-rigid objects~\cite{zhang2021cross,shen2021much}; CLIP-ViT, while offering global context modeling, incurs high computational cost due to quadratic self-attention and struggles to capture fine-grained structures under fixed patch partitioning, restricting its ability to represent local details and deformations. These limitations highlight the need for a visual framework that combines efficient global modeling with dynamic features for robust language-guided grasping.

Accurate spatial modeling is crucial in cluttered or occluded scenes, where objects overlap or exhibit substantial visual similarity. Recent studies have introduced depth information to enhance grasping by providing structural cues such as object shape, spatial relationships, and occlusion reasoning~\cite{lin2025prompting,yu2025co}.
However, depth maps from real-world sensors are often noisy or incomplete due to reflections, translucent materials, or missing regions. Existing RGB-D methods typically adopt dual-encoder architectures that process RGB and depth separately. While this extracts complementary features, it also incurs computational redundancy.
Furthermore, low-quality or inconsistent depth data can hinder RGB feature learning~\cite{bai2025v,bodenstedt2019prediction,li2025roboflamingo}, weakening spatial reasoning and target discrimination. Thus, it is essential to design strategies that use depth as geometric guidance while preserving RGB integrity and avoiding redundant computation~\cite{marin2021token}.

Beyond deep guidance, effective language-guided grasping also hinges on precise cross-modal alignment that unifies semantic understanding with spatial reasoning. While language offers semantic cues for target identification and depth provides structural guidance, associating textual descriptions with corresponding visual regions remains highly challenging in cluttered or occluded environments~\cite{yi2024text}. In prevailing approaches, the vision encoder is typically frozen and only high-level features are passed to the connector to reduce computational cost, leaving informative low- and mid-level representations underutilized. Moreover, common cross-modal fusion strategies, such as concatenation or cross-attention, fail to fully exploit multi-layer visual features~\cite{tang2025geo}, thereby diluting task-relevant signals and overlooking critical semantic or spatial cues~\cite{chen2023perception}. Approaches that either select a subset of visual layers or concatenate all layers further exacerbate this issue, as they produce high-dimensional representations that increase computational overhead and complicate optimization~\cite{zhang2021cross}. Collectively, these limitations hinder precise text-to-vision alignment in complex grasping scenarios, ultimately constraining the discriminative power and generalization ability of multimodal representations.

To address these challenges, we introduce GeoLanG (Geometry-Aware Language-Guided Grasping), a unified multi-task visual-language framework to perform reliable grasping in open-world environment. Our contributions are summarized as follows:

\begin{itemize}
    \item We present GeoLanG, an end-to-end framework that integrates RGB-D perception with natural language understanding into a shared representation space, enabling robust semantic alignment and improved generalization in cluttered and complex environments.

    \item To enhance spatial reasoning, we introduce the Depth-guided Geometric Module (DGGM), which encodes depth as explicit geometric priors directly into the attention mechanism, improving robustness against occlusions and target discrimination without the need for a dedicated depth encoder.

    \item To facilitate effective cross-modal fusion, we propose Adaptive Dense Channel Integration (ADCI), which adaptively aggregates multi-layer visual features and emphasizes task-relevant signals, yielding more discriminative and generalizable visual representations.

    \item Extensive experiments on the OCID-VLG dataset demonstrate that GeoLanG surpasses existing methods in both language segmentation and grasping metrics, while maintaining robustness under occlusions and ambiguous instructions, validating the effectiveness of our unified multi-task approach.
\end{itemize}

\section{Method}
\subsection{Overall Architecture of GeoLanG}
\begin{figure*}[!t]
\centering
\includegraphics[width=0.9\textwidth]{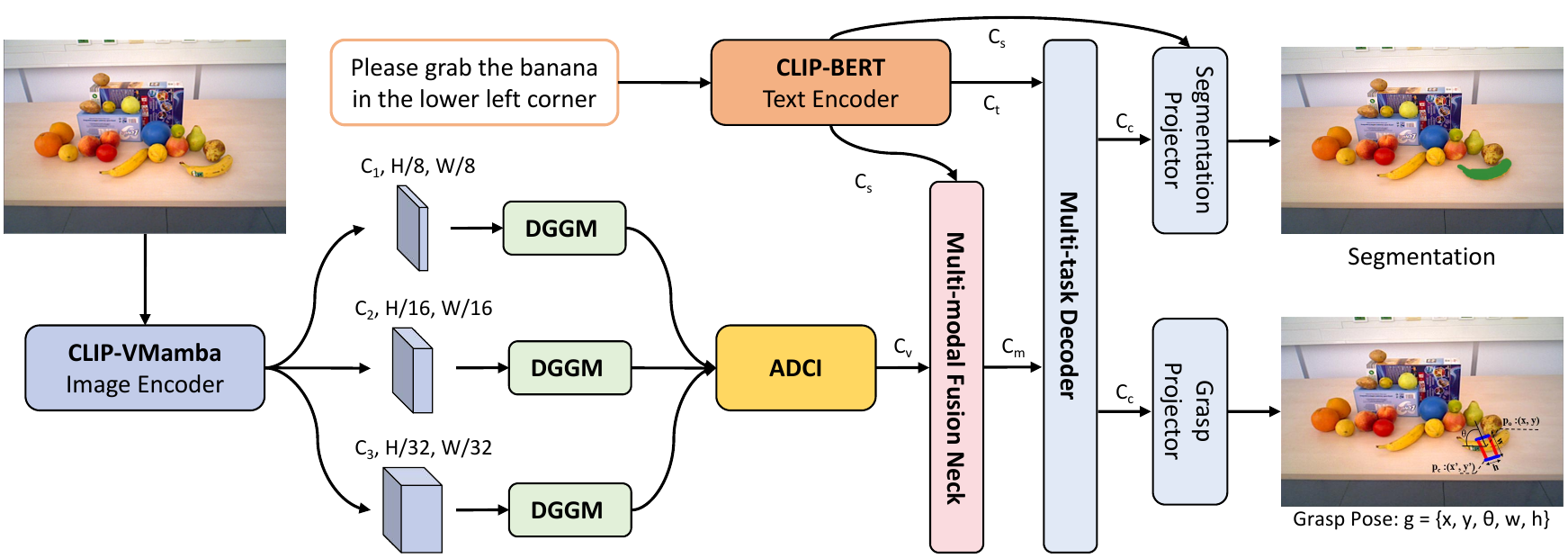}
\caption{An overview of the GeoLanG framework. Given an RGB-D image and a language query, the text encoder extracts high-level semantic features while the RGB encoder processes multi-scale visual information. Depth-derived geometric priors are incorporated via the Depth-Guided Geometric Module (DGGM) to enhance structural cues. Multi-layer visual features are then optimized through ADCI before being fed into dual-path projectors that generate pixel-level segmentation masks and refine the grasp pose for the target object.}
\label{fig_method}
\end{figure*}

As shown in Fig.~\ref{fig_method}, GeoLanG is an end-to-end vision-language grasping framework that is specifically designed to enable precise language-guided grasping in complex open scenes. In this framework, given an RGB-D image and a language query, the text encoder first extracts high-level semantic features, while simultaneously, the RGB encoder processes multi-scale visual information. 

We follow the CLIP paradigm to align visual and linguistic features into a shared semantic space. For the image encoder, existing CLIP visual backbones face notable challenges in complex grasping scenarios. CLIP-ResNet relies on a rigid hierarchical design that restricts multi-scale adaptability, while CLIP-ViT conducts self-attention over fixed-size patches, leading to high computational cost and limited sensitivity to fine-grained spatial cues. Therefore, we employ CLIP-VMamba~\cite{huang2024clip}, which integrates the complementary advantages of ViT and CNN architectures: the ViT backbone captures long-range dependencies and global context, whereas the CNN-style modules preserve high-resolution local details. This hybrid design effectively represents non-rigid and multi-scale objects in cluttered open scenes. In the CLIP-VMamba visual encoder, we retain the last three scales of features, excluding the lowest-level feature to balance computational cost and semantic richness. Formally, the retained multi-scale features can be denoted as $C = \{C_i \in \mathbb{R}^{H/s_i \times W/s_i \times C_i} \mid i=1,2,3\}$, where $C_1$ preserves spatial details for precise localization, and $C_2$ and $C_3$ encode higher-level semantic concepts. We set the downsampling factors to $s_1=8$, $s_2=16$, and $s_3=32$, respectively. Then, the Depth-Guided Geometric Module (DGGM) incorporates depth-derived geometric priors to enhance structural cues. 
These spatial-enhanced features are optimized through Adaptive Dense Channel Integration (ADCI) to produce multi-scale visual representations $C_v \in \mathbb{R}^{N \times C}$, where $N = H/16 \cdot W/16$. Meanwhile, for language inputs, the textual instructions $P = \{w_1, \dots, w_T\}$ are encoded with CLIP-BERT~\cite{lei2021less} to obtain token embeddings $C_t \in \mathbb{R}^{K \times C}$ and a sentence embedding $C_s \in \mathbb{R}^{C'}$, capturing both local and global textual semantics.

In multi-task decoding, $C_v$ is fused with the sentence embedding $C_s$ through the Multimodal Fusion Neck to produce pixel-wise multimodal representations $C_m \in \mathbb{R}^{N \times C}$, which are subsequently fed into the downstream multi-task decoder with $C_t$  to produce $C_c \in \mathbb{R}^{N \times C}$. In the segmentation branch, $C_c$ and $C_s$ are projected into a shared space and supervised via a binary cross-entropy loss applied to the dot product of the projected embeddings. The final segmentation mask is obtained by reshaping and upsampling the projected features, while the grasp head predicts grasp configurations from the same shared features, enabling joint optimization of perception and action reasoning.

\subsection{Depth-guided Geometric Module}

\begin{figure}[!t]
\centering
\includegraphics[width=0.45\textwidth]{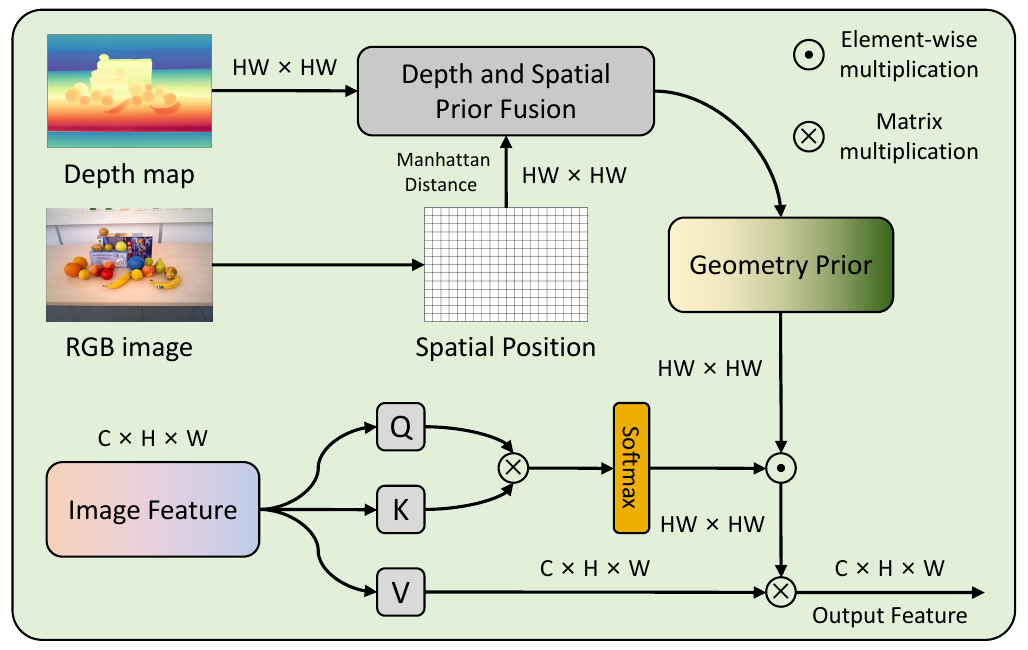}

\caption{The overview of the Depth-guided Geometric Module (DGGM). The pink rectangle represents image features extracted from the RGB encoder, the green rectangle denotes the learned geometry prior, and the yellow rectangle indicates the spatial prior computed from depth and RGB. $\otimes$ represents matrix multiplication and $\odot$ represents element-wise multiplication. The output feature integrates visual information with geometric and spatial cues to enhance multi-scale representations.}

\label{method_2}
\end{figure}

To effectively incorporate depth cues while avoiding excessive computational overhead, we introduce depth-derived geometric priors~\cite{yin2025dformerv2} and fuse them with RGB features from the V-Mamba encoder (Fig.~\ref{method_2}). Specifically, a 2D input RGB image of size $h \times w$ is evenly partitioned into $H \times W$ patches, where $H$ and $W$ denote the number of patches along rows and columns, respectively. Each patch $Q_{m,n}$ is indexed by its spatial coordinates $(m,n)$. When the corresponding depth map is available, the patch at the same position encodes its relative distance from the camera plane. By integrating these two modalities, we explicitly model the geometric relationships among all patches and embed them into the self-attention mechanism, capable of adapting to multi-scale image features.

For the depth prior, we apply average pooling to all pixels in the depth patch at $(m,n)$ to obtain its representative depth value $D_{m,n}$. The depth difference between any two patches is defined as:
\begin{equation}
\Delta D_{m,n,m',n'} = | D_{m,n} - D_{m',n'} |
\end{equation}
where $\Delta D \in \mathbb{R}^{HW \times HW}$ forms the depth relationship matrix. The spatial distance is calculated through the Manhattan distance between patch coordinates:
\begin{equation}
\Delta S_{m,n,m',n'} = | m - m' | + | n - n' |
\end{equation}
where $\Delta S \in \mathbb{R}^{HW \times HW}$ is the spatial relationship matrix. To unify depth and spatial priors, the fused geometry prior $\mathcal{G}$ is constructed by:
\begin{equation}
\mathcal{G} = \lambda_1 \cdot \Delta D + \lambda_2 \cdot \Delta S
\end{equation}
Where $\lambda_1, \lambda_2$ are learnable parameters balancing the contributions of depth and spatial priors, the resulting $\mathcal{G} \in \mathbb{R}^{HW \times HW}$ encodes comprehensive 3D geometric relationships among all patches.

For the integration of the geometry prior $\mathcal{G}$ into the attention mechanism, given an RGB feature map $X \in \mathbb{R}^{HW \times C}$, we can obtain the query, key, and value matrices by linear projections:$Q, K, V \in \mathbb{R}^{HW \times d}$. We integrate the geometry prior $\mathcal{G}$ via a decay factor $\eta \in (0,1)$ to emphasize nearby patches:
\begin{align}
\hat{X} &= \big( \text{Softmax}(Q K^\top) \odot \eta \, \mathcal{G} \big)V^\top
\end{align}
Smaller values in $\eta \mathcal{G}$ correspond to larger geometric distances, suppressing irrelevant key-value pairs and emphasizing geometrically relevant ones.

This formulation can be extended to multi-head attention by assigning different decay rates $\eta$ for each head. Through DGGM, our model effectively captures spatial dependencies, object shape variations, and low-texture regions, while maintaining computational efficiency.

\subsection{Adaptive Dense Channel Integration}


Prevailing paradigms in visual representation learning often freeze backbones and only propagate high-level semantic features, which discards valuable low- and mid-level cues. In this work, we propose the Adaptive Dense Channel Integration (ADCI) module to holistically harness this multi-layer information. ADCI first preserves dense cross-layer connections while adaptively balancing contributions from different layers. Given $L$ image features $\{C_1, \dots, C_L\}$ extracted from V-Mamba, they are divided into $G$ groups, each containing $M = L/G$ consecutive layers. Learnable weights $\alpha_i$ within each group adaptively adjust layer contributions based on input features, enabling dynamic information aggregation. For the $g$-th group of $M$ consecutive layers, the group-level feature $GC_g$ is computed as a weighted combination of its constituent features:
\begin{equation}
GC_g = \sum_{i=(g-1)M+1}^{gM} \alpha_i C_i, \quad 1 \le g \le G
\end{equation}
where the adaptive weights $\alpha_i$ are predicted by a lightweight gating network and normalized to satisfy
\begin{equation}
\sum_{i=(g-1)M+1}^{gM} \alpha_i = 1
\end{equation}

The gating network first summarizes each feature map $C_i$ into a compact descriptor $z_i$ using global average pooling (GAP):
\begin{equation}
z_i = \text{GAP}(C_i)
\end{equation}
Which is then passed through a lightweight two-layer MLP to produce unnormalized scores $s_i$:
\begin{equation}
s_i = W_2 \,\sigma(W_1 z_i + b_1) + b_2
\end{equation}
Where $W_1$ and $W_2$ are learnable parameters and $\sigma$ denotes a non-linear activation function (e.g., ReLU). Softmax normalization within each group yields the adaptive weights $\alpha_i$:
\begin{equation}
\alpha_i = \frac{\exp(s_i)}{\sum_{j=(g-1)M+1}^{gM} \exp(s_j)}
\end{equation}

Finally, the group-level features $\{GC_1, \dots, GC_G\}$ are concatenated with the last-layer feature $C_L$ and projected through an MLP to generate the final visual embedding:
\begin{align}
e_v = \text{MLP} \Big( \text{Concatenate}([GC_1, \dots, GC_G, C_L], \notag\\
\text{dim=channel})\Big)
\end{align}

This formulation allows ADCI to balance contributions across layers while preserving dense connections, ensuring informative low- and mid-level features are effectively aggregated. Consequently, it yields richer visual representations that enhance cross-modal expressiveness for downstream tasks while maintaining efficiency. Algorithm~\ref{alg:adci_mamba} shows the flowchart for ADCI.

\begin{algorithm}[t]
\caption{Adaptive Dense Channel Integration (ADCI) for Mamba Features}
\label{alg:adci_mamba}
\KwIn{Multi-layer visual features $\{C_1, \dots, C_L\}$ from VMamba image encoder}
\KwOut{Final visual embedding $e_v$}

\BlankLine
\textbf{Step 1: Grouping.} Partition $\{C_1, \dots, C_L\}$ into $G$ groups, each with $M = L/G$ layers. \\

\For{$g \gets 1$ \textbf{to} $G$}{
    \BlankLine
    \textbf{Step 2: Feature descriptors.} For each feature $C_i$ in group $g$: \\
    \Indp $z_i = \text{GAP}(C_i)$ \tcp*{Global average pooling} \Indm
    
    \textbf{Step 3: Gating network.} Compute score via 2-layer MLP: \\
    \Indp $s_i = W_2 \,\sigma(W_1 z_i + b_1) + b_2$ \Indm
    
    \textbf{Step 4: Adaptive weights.} Apply softmax within the group: \\
    \Indp $\alpha_i = \dfrac{\exp(s_i)}{\sum_{j=(g-1)M+1}^{gM} \exp(s_j)}$ \Indm
    
    \textbf{Step 5: Group feature aggregation.} \\
    \Indp $GC_g = \sum_{i=(g-1)M+1}^{gM} \alpha_i C_i$
}

\BlankLine
\textbf{Step 6: Final embedding.} Concatenate group features with last layer $C_L$: \\
\Indp $e_v = \text{MLP}\Bigl(\text{Concatenate}([GC_1, \dots, GC_G, C_L],$ \\
      \makebox[0.8\linewidth][r]{$\text{dim=channel})\Bigr)$}

\end{algorithm}

\section{Experiment}
\subsection{Dataset and Experiment setup}
\begin{figure}[h]
    \centering               
\includegraphics[width=0.48\textwidth]{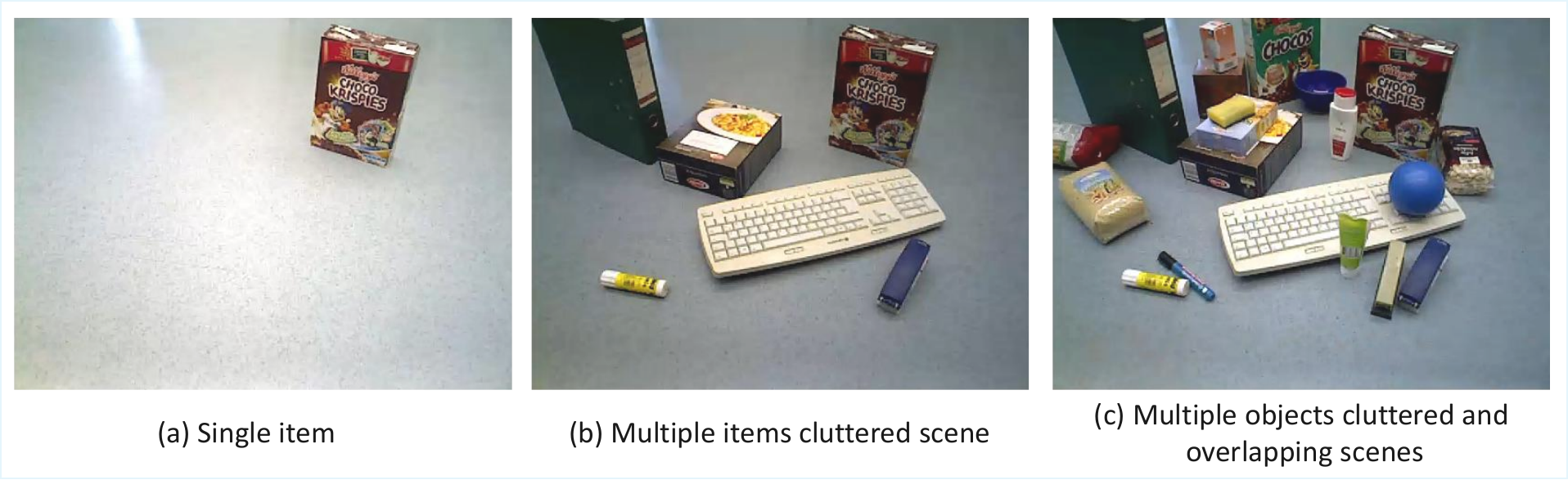}
    \caption{Examples from the OCID-VLG dataset~\cite{tziafas2023language}. (a)Single item (b) Multiple items cluttered scene (c) Multiple objects cluttered and overlapping scenes.} 
    \label{fig_dataset}
\end{figure}

To evaluate our approach on language-guided segmentation and grasping, we use the OCID-VLG benchmark~\cite{tziafas2023language}. As shown in Fig.~\ref{fig_dataset}, OCID-VLG extends the OCID cluttered indoor scenes with multimodal annotations, bridging language supervision and robotic grasping. It contains 1,763 RGB-D tabletop scenes across 31 categories and 58 fine-grained instances, with labels covering semantic attributes, spatial relations, 2D/3D locations, and 4-DoF grasps. By incorporating CLEVR-style data generation, the dataset provides 89,639 triplets of language expressions, segmentation masks/bounding boxes, and grasp annotations, enabling unified multimodal supervision.

\subsubsection{Experimental Setup} We initialize the visual and textual encoders with pre-trained CLIP-VMamba and BERT weights, respectively. The resolution of input images is $416 \times 416$, while text sequences are tokenized using byte-pair encoding (BPE)~\cite{vaswani2017attention,radford2019language} with a maximum length of 20 tokens. Training is performed on four NVIDIA RTX 4090 GPUs using the Adam optimizer for 50 epochs, with an initial learning rate of $1 \times 10^{-4.5}$ and a batch size of 64.

\begin{table*}[h]
\vspace{0.2cm} 
\centering
\caption{Comparison experiments against SOTA solutions on the OCID-VLG dataset.}
\setlength{\tabcolsep}{8pt} 
\begin{tabular}{l|cccccc|cc}
\hline
\multirow{2}{*}{Method} & \multicolumn{6}{c|}{Segmentation Task} & \multicolumn{2}{c}{Grasping Task} \\ \cline{2-9}
 & IoU & Pr@50 & Pr@60 & Pr@70 & Pr@80 & Pr@90 & J@1 & J@N \\ \hline
VLG            & 76.35 & 89.84 & 85.38 & 77.63 & 59.67 & 20.36 & 78.51 & 85.19 \\
GraspCLIP      & 77.20 & 91.34 & 87.85 & 80.88 & 62.52 & 13.50 & 78.32 & 84.73 \\
CLIPort        & 78.11 & 94.03 & 90.84 & 82.73 & 60.08 & 15.68 & 83.90 & 88.75 \\
CTNet          & 80.94 & \textbf{97.45} & 95.43 & 89.63 & 67.05 & 16.66 & 84.32 & 90.91 \\
CROG           & 80.77 & 95.87 & 93.83 & 86.69 & 65.14 & \textbf{16.85} & 81.64 & 88.05 \\
GeoLanG (Ours) & \textbf{85.77} & 96.73 & \textbf{95.45} & \textbf{89.82} & \textbf{68.23} & 16.70 & \textbf{87.32} & \textbf{92.13} \\ \hline
\end{tabular}
\label{tab:table_1}
\end{table*}

\begin{figure*}[h]
    \centering               
\includegraphics[width=1.0\textwidth]{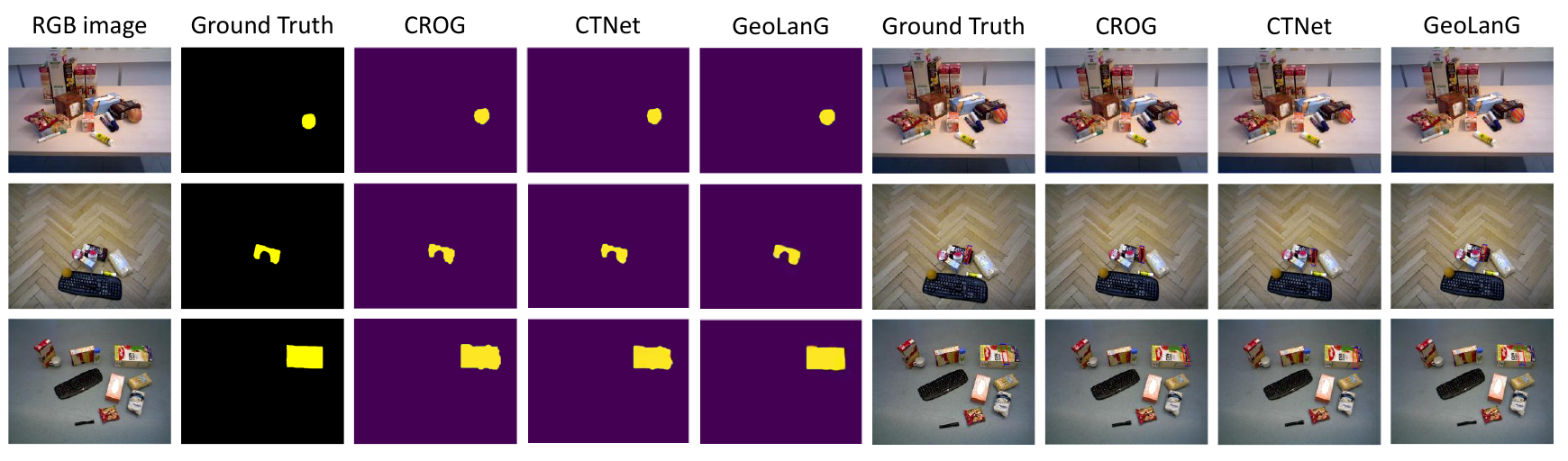}
    \caption{Qualitative comparison against SOTA solutions on the OCID-VLG dataset.}
    \label{fig_visual}
\end{figure*}

\subsubsection{Evaluation Metrics} For referring image segmentation (RIS)~\cite{wang2022cris,ding2021vision}, we adopt Intersection-over-Union (IoU) and Precision@\textit{X}. IoU measures the average overlap between predicted masks and ground-truth annotations, while Precision@\textit{X} computes the percentage of predictions whose IoU exceeds thresholds $X \in \{0.5, 0.6, 0.7, 0.8, 0.9\}$. For referring grasp synthesis (RGS), we report the Jacquard index J@\textit{N}~\cite{depierre2018jacquard,xu2023instance,ainetter2021end}, which evaluates the proportion of the top-$N$ grasp predictions that achieve an intersection-over-union greater than $0.25$ with the ground-truth grasp rectangle and an orientation difference within $30^{\circ}$.

\subsection{Comparison with baselines on OCID-VLG}
We first evaluate our proposed GeoLanG framework on the OCID-VLG dataset and compare it with several CLIP-based baselines, including CROG~\cite{tziafas2023language}, GraspCLIP~\cite{tang2023task}, CLIPort~\cite{shridhar2022cliport}, and CTNet~\cite{xie2024listen}. As shown in Table~\ref{tab:table_1}, GeoLanG achieves superior performance on both segmentation and grasping metrics. For segmentation, the model attains an IoU of 85.77\%, substantially outperforming CTNet (80.94\%) and CROG (80.77\%), demonstrating its ability to produce more accurate pixel-level object masks. Under stricter evaluation, GeoLanG maintains consistent robustness, achieving 89.82\% at Pr@70 compared with 89.63\% for CTNet and 86.69\% for CROG, indicating effective multi-scale feature aggregation and precise localization even under challenging conditions. Although its performance at Pr@90 (16.70\%) is slightly lower than CROG (16.85\%), the overall stability across varying thresholds highlights the model's balanced precision-recall trade-off. In the grasping task, GeoLanG reaches the highest performance with J@1 of 87.32\% and J@N of 92.13\%, surpassing CTNet by 3.00 and 1.22 points, respectively. CLIPort achieves competitive grasp metrics (J@1 = 83.90\%, J@N = 88.75\%) but underperforms, primarily due to coarser pixel-level segmentation and weaker depth-aware cues, which constrain reliable grasp prediction in cluttered or occluded scenes. These results demonstrate that integrating RGB-D priors with dense multi-scale feature fusion and depth-guided geometric reasoning enables GeoLanG to achieve both accurate segmentation and grasp prediction, yielding strong generalization in complex environments.

Since CROG and CTNet address the same task as our proposed GeoLanG and incorporate attribute-aware segmentation, we further conduct a qualitative comparison. As illustrated in Fig.~\ref{fig_visual}, GeoLanG delivers more precise fine-grained recognition, generating segmentation masks that closely follow object boundaries. This advantage is especially evident in challenging cases such as small objects, blurry edges, and occlusions (e.g., first and second rows). In contrast, CROG and CTNet often yield oversimplified masks (e.g., retaining only central regions) or misaligned boundaries, indicating weaker semantic alignment under complex visual conditions. For grasp pose prediction, CROG and CTNet often display grasp drift or incorrectly target neighboring objects, while  GeoLanG consistently localizes grasp points on small objects within cluttered scenes. These advantages stem from GeoLanG's enhanced spatial reasoning and depth-aware geometric perception, which strengthen its robustness in cluttered and occluded environments.

\begin{table*}[h]
\vspace{0.2cm} 
\centering
\caption{Comparison experiments against SOTA solutions on the \textit{novel instance} split of OCID-VLG.}
\setlength{\tabcolsep}{8pt} 
\begin{tabular}{l|cccccc|cc}
\hline
\multirow{2}{*}{Method} & \multicolumn{6}{c|}{Segmentation Task} & \multicolumn{2}{c}{Grasping Task} \\ \cline{2-9} 
 & IoU & Pr@50 & Pr@60 & Pr@70 & Pr@80 & Pr@90 & J@1 & J@N\\ \hline
VLG            & 69.70 & 77.03 & 70.16 & 63.31 & 49.58 & 14.78 & 66.42 & 74.91 \\
GraspCLIP      & 69.13 & 83.21 & 76.44 & 63.80 & 42.20 & 8.43  & 63.98 & 58.16 \\
CLIPort        & 71.17 & 80.51 & 74.23 & 67.75 & 53.77 & 18.55 & 66.24 & 74.03 \\
CTNet          & 74.38 & 87.34 & 81.63 & 73.89 & 57.23 & 17.83 & 72.56 & 79.06 \\
CROG           & 73.28 & 82.98 & 72.22 & 81.41 & 56.61 & \textbf{20.20} & 71.36 & 81.41 \\
GeoLanG (Ours) & \textbf{81.25} & \textbf{96.32} & \textbf{93.48} & \textbf{88.19} & \textbf{63.78} & 15.42 & \textbf{83.96} & \textbf{90.72} \\ \hline
\end{tabular}
\label{tab:table_2}
\end{table*}

\begin{table*}[h]
\centering
\caption{Ablation study on Image Encoder, Depth-guided Geometric Module and Adaptive Dense Channel Integration.}
\setlength{\tabcolsep}{6pt} 
\begin{tabular}{ccc|cccccc|cc}
\hline
\multirow{2}{*}{CLIP-VMamba} & \multirow{2}{*}{DGGM} & \multirow{2}{*}{ADCI} & \multicolumn{6}{c|}{Segmentation Task} & \multicolumn{2}{c}{Grasping Task} \\ \cline{4-11}
 & & & IoU & Pr@50 & Pr@60 & Pr@70 & Pr@80 & Pr@90 & J@1 & J@N \\ \hline
\scriptsize{\XSolidBrush} & \scriptsize{\XSolidBrush} & \scriptsize{\XSolidBrush} & 80.77 & 95.87 & 93.83 & 86.69 & 65.14 & 16.85 & 81.64 & 88.05 \\
\scriptsize{\CheckmarkBold} & \scriptsize{\XSolidBrush} & \scriptsize{\XSolidBrush} & 83.25 & 96.23 & 94.39 & 86.93 & 67.04 & 16.91 & 83.44 & 90.76 \\
\scriptsize{\CheckmarkBold} & \scriptsize{\CheckmarkBold} & \scriptsize{\XSolidBrush} & 84.88 & 96.46 & 95.27 & 88.38 & 67.52 & \textbf{17.02} & 85.35 & 91.28 \\
\scriptsize{\CheckmarkBold} & \scriptsize{\CheckmarkBold} & \scriptsize{\CheckmarkBold} & \textbf{85.77} & \textbf{96.73} & \textbf{94.45} & \textbf{89.82} & \textbf{68.23} & 16.70 & \textbf{87.32} & \textbf{92.13} \\ \hline
\end{tabular}
\label{tab:table_3}
\end{table*}

\subsection{Comparison with baselines on unseen items (OCID-VLG)}

To evaluate generalization, we comprehensively test GeoLanG on the \textit{novel instance} split of OCID-VLG (Table~\ref{tab:table_2}). GeoLanG achieves the highest overall performance, with IoU 81.25\%, J@1 83.96\%, and J@N 90.72\%, clearly surpassing CTNet by 6.87, 11.66, and 11.66 points on IoU, J@1, and J@N, respectively. In addition, GeoLanG achieves strong and stable segmentation at mid thresholds (Pr@50 = 96.32\%, Pr@60 = 93.48\%, Pr@70 = 88.19\%) while still outperforming baselines in grasping, despite slightly lower Pr@90 (15.42\%) than CROG (20.20\%). Overall, GeoLanG demonstrates robust and consistent segmentation-grasping predictions on novel objects. These performances primarily stem from the Depth-guided Geometric Module, which effectively uses RGB-D priors to maintain segmentation and grasping accuracy for occluded or unseen objects, and the Adaptive Dense Channel Integration, which adaptively aggregates multi-layer features for precise localization.

\subsection{Ablation Studies}

To validate the effectiveness of our proposed components in GeoLanG, we conduct an ablation study on the CLIP-VMamba backbone, DGGM, and ADCI, with results summarized in Table~\ref{tab:table_3}. When using CLIP-ResNet instead of CLIP-VMamba, the model struggles to capture sufficient spatial reasoning and cross-modal semantics, leading to significantly lower grasping and segmentation performance (J@1 = 81.64\%, J@N = 88.05\%, IoU = 80.77\%). Incorporating CLIP-VMamba alleviates this issue, as its stronger representation learning boosts performance across all metrics, confirming its importance for multi-scale visual-textual alignment. Building on this, the incorporation of DGGM enables explicit modeling of RGB-D geometric priors, leading to notable gains in structural sensitivity and robustness, as reflected by higher Pr@90 (17.02\%). Further, the integration of ADCI strengthens multi-layer feature aggregation, allowing the model to capture fine-grained localization cues and ultimately achieve the best overall trade off across segmentation and grasping tasks, particularly at higher precision thresholds (e.g., Pr@80) (J@1 = 87.32\%, J@N = 92.13\%, IoU = 85.77\%, Pr@80 = 68.23\%). These results demonstrate that each module provides complementary benefits CLIP VMamba enhances representation capacity, DGGM enriches geometric reasoning, and ADCI facilitates dense feature fusion—together driving robust and high-precision referring grasping in challenging scenarios.

\begin{figure}[h]
    \centering               
\includegraphics[width=0.48\textwidth]{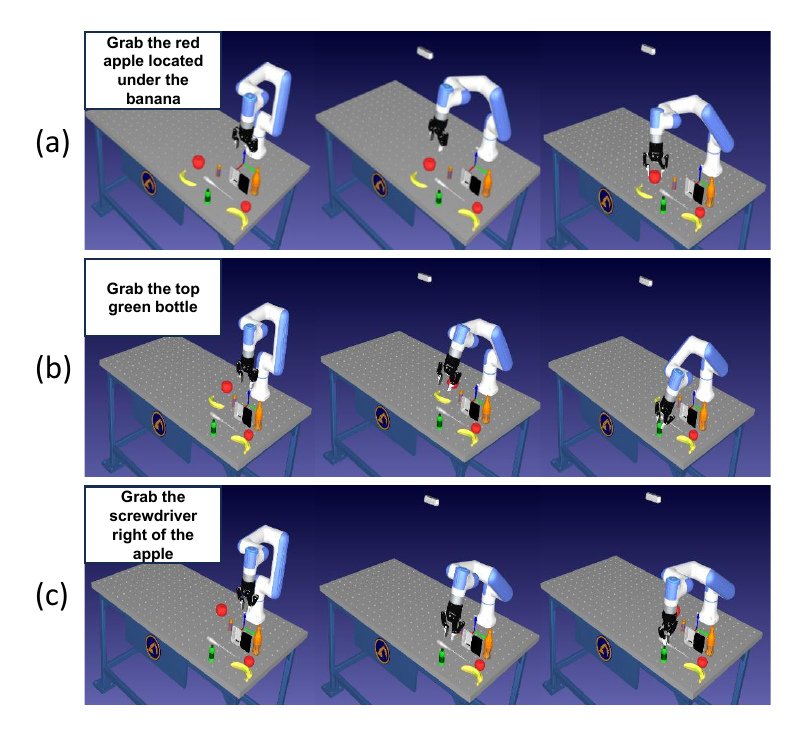}
    \caption{Simulation experiments on the RoboDK platform.} 
    \label{fig_sim}
\end{figure}

\begin{figure*}[!t]
\centering
\includegraphics[width=0.95\textwidth]{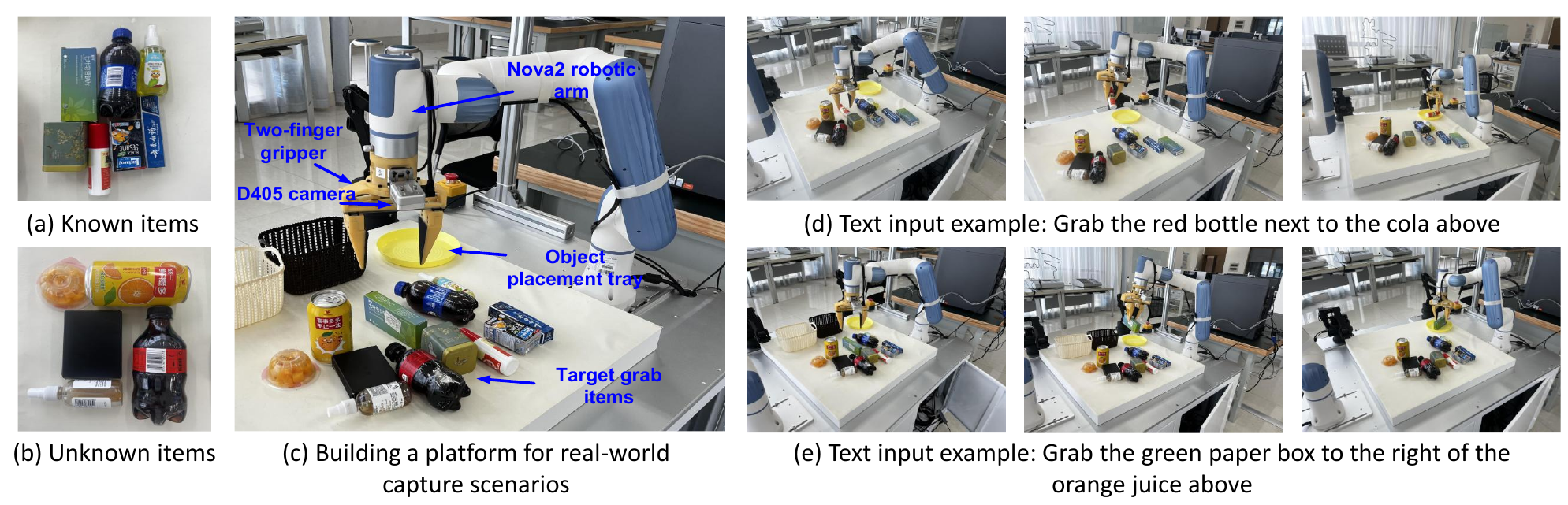}
\caption{Real-world experiments on the DOBOT Nova 2 robo.}
\label{fig_robot}
\end{figure*}

\begin{table*}[h]
\setlength{\tabcolsep}{6pt}
\centering    
\footnotesize
\caption{Simulation and real robot experimental results}
\begin{tabular}{c|cccccccc|c}
\hline
Setting       & Toothbrush    & Mug        & Fruit        & Bottle       & Spray      & Medicine   & Snack Box   & Total       & Sim \\ \hline
Isolated Scene (Count) & 3           & 5         & 7           & 3           & 2         & 10        & 10         & 40          & 40   \\ \cline{1-1}
Segmentation IOU & 3(100\%)    & 4 (80\%)  & 6 (85.71\%) & 3 (100\%)   & 2 (100\%) & 6 (60\%)  & 10(100\%)  & 34 (85\%)   & 98.21\%   \\ \cline{1-1}
Grasping Accuracy     & 3(100\%)    & 4 (80\%)  & 6 (85.71\%) & 2 (66.67\%) & 2 (100\%) & 5 (50\%)  & 10(100\%)  & 33 (82.5\%) & 100\% \\ \hline
Cluttered Scene (Count) & 3           & 5         & 7           & 3           & 2         & 10        & 10         & 40          & 40 \\ \cline{1-1}
Segmentation IOU & 3(100\%)    & 5 (100\%) & 6 (85.71\%) & 2 (66.67\%) & 2 (100\%) & 6 (60\%)  & 9 (90\%)   & 32 (80\%)   & 94.21\%    \\ \cline{1-1}
Grasping Accuracy     & 2 (66.67\%) & 4 (80\%)  & 4 (57.14\%) & 1 (33.33\%) & 1 (50\%)  & 4 (40\%)  & 8 (80\%)   & 24 (60\%)   & 92\% \\ \hline
\end{tabular}
\label{tab:table_4}
\vspace{-3mm}
\end{table*}

\subsection{Robot Experiments}
We further evaluate GeoLanG on a table-cleaning task through both simulation and real-world experiments. To mitigate the observation domain gap between CLIP pretraining and GeoLanG training on the OCID-VLG dataset, we fine-tune the vision-language encoders end-to-end on the target data and incorporate depth-guided geometric features, which reduce sensitivity to appearance variations and improve cross-domain robustness. The simulation study is conducted on the RoboDK platform using an eye-to-hand configuration for efficient setup (Fig.~\ref{fig_sim}), whereas the real-world experiment employs a DOBOT Nova 2 robot equipped with parallel-jaw grippers and an Intel RealSense D405 camera in an eye-in-hand configuration (Fig.~\ref{fig_robot}(c)), enabling accurate grasp execution. The eye-in-hand configuration is adopted for practical deployment and accurate viewpoint alignment during grasp execution, while the perception and grasp prediction are performed in a single-shot manner without closed-loop visual feedback. Example trials in both isolated and cluttered scenes are shown in Fig.~\ref{fig_robot}(d) and (e). Notably, although the robotic arm offers 6-DoF motion, 4-DoF grasp annotations are sufficient, as the remaining degrees of freedom primarily influence approach orientation rather than grasp location or width.

We design two experimental settings to evaluate GeoLanG: \textit{isolated} scenes, where objects are spatially separated, and \textit{cluttered} scenes, where objects are densely packed. To further increase the challenge, each scene includes distractors from the same category as the query object (Fig.~\ref{fig_robot}(a)(b)). The evaluation covers eight object categories, consisting of four seen classes (toothbrush, mug, fruit, beverage bottle) and three unseen classes (spray, medicine, snack box). In total, 40 scenes are constructed (20 isolated, 20 cluttered) with each scene paired with a corresponding language instruction.

As shown in Table~\ref{tab:table_4}, GeoLanG achieves high performance in simulation, with grasping accuracy reaching 95\%  across all categories. In real-world isolated scenes, segmentation and grasping accuracies are 85\% and 82.5\%, respectively, while in cluttered scenes they decrease to 80\%  and 60\%, reflecting the impact of sensor noise, calibration errors, and occlusions. Detailed per-category results show that objects with minimal occlusion, such as toothbrushes and spray bottles, achieve near-perfect segmentation and grasping, whereas heavily occluded or small objects, such as medicine boxes, exhibit lower performance. Despite unseen objects and complex occlusions, GeoLanG consistently localizes attribute concepts and uses spatial relationships to distinguish target objects. Overall, the results highlight GeoLanG’s strong generalization to novel objects, robustness under both virtual and real-world imperfections, and effectiveness in executing table-cleaning tasks across both structured and cluttered environments.

\section{Dicussion}
GeoLanG demonstrates competitive performance on language-guided grasping under standard benchmark settings. The current framework, however, adopts a simplified formulation by focusing on 4-DoF planar grasping, which reduces the complexity of gripper orientation and interaction dynamics. While this design is consistent with common practice in existing benchmarks, it constrains the range of manipulation scenarios. The evaluation protocol reflects this formulation. Experiments are primarily conducted on the OCID-VLG benchmark and set of real-world table-cleaning tasks, offering a controlled but relatively narrow coverage of real-world variability. Consequently, the behavior of the model in more diverse or dynamic environments has yet to be fully characterized.
These considerations point to several directions for future work, including extending the framework to full 6-DoF grasping and validating its performance in broader and more dynamic real-world settings. Such extensions would enable a more comprehensive assessment of the generality of GeoLanG for complex language-conditioned manipulation tasks.

\section{Conclusions}
This work introduces GeoLanG, an end-to-end vision-language grasping framework that integrates RGB-D perception with natural language understanding. GeoLanG adopts the CLIP paradigm, employing a CLIP-VMamba visual encoder and a CLIP-BERT text encoder to establish a shared representation space, thereby strengthening cross-modal alignment and improving semantic discrimination. To make full use of depth cues, the Depth-Guided Geometric Module injects geometric priors into RGB features, strengthening spatial awareness and improving robustness under occlusion and low-texture conditions. Besides, the Adaptive Dense Channel Integration module further integrates multi-layer features, boosting the discriminative power of visual representations. Extensive experiments demonstrate the effectiveness of our proposed methods in challenging scenarios, showing improved target localization, enhanced object understanding, and efficient representation learning for downstream tasks including grasp detection, segmentation, and instruction parsing.





\bibliographystyle{IEEEtran}
\bibliography{icra}

\end{document}